\documentclass[sn-basic,Namedate]{sn-jnl}

\usepackage{graphicx}%
\usepackage{multirow}%
\usepackage{amsmath,amssymb,amsfonts}%
\usepackage{amsthm}%
\usepackage{mathrsfs}%
\usepackage[title]{appendix}%
\usepackage{xcolor}%
\usepackage{textcomp}%
\usepackage{manyfoot}%
\usepackage{booktabs}%
\usepackage{algorithm}%
\usepackage{algorithmicx}%
\usepackage{algpseudocode}%
\usepackage{listings}%

\begin{document}

\title[Article Title]{Enhancing actuarial non-life pricing models via transformers}

\author[1]{\fnm{Alexej} \sur{Brauer} \orcid{https://orcid.org/0009-0009-4001-016X}}\email{brauer.alexej@gmail.com}

\affil[1]{\orgdiv{Actuarial Department}, \orgname{Allianz Versicherungs-AG}, \orgaddress{\street{Königinstr. 28}, \city{Munich}, \postcode{80802}, \state{Bavaria}, \country{Germany}}}

\abstract{
Currently, there is a lot of research in the field of neural networks for non-life insurance pricing.
The usual goal is to improve the predictive power via neural networks 
while building upon the generalized linear model, which is the current industry standard. 
Our paper contributes to this current journey via novel methods 
to enhance actuarial non-life models with transformer models for tabular data. 
We build here upon the foundation laid out by the combined actuarial neural network as well as the localGLMnet 
and enhance those models via the feature tokenizer transformer.
The manuscript demonstrates the performance of the proposed methods on a real-world claim frequency dataset and 
compares them with several benchmark models such as generalized linear models, feed-forward neural networks, combined actuarial neural networks, 
LocalGLMnet, and pure feature tokenizer transformer.
The paper shows that the new methods can achieve better results than the benchmark models 
while preserving certain generalized linear model advantages. The paper also discusses the practical implications and 
challenges of applying transformer models in actuarial settings.
}
\keywords{non-life insurance, transformer, tabular data, regression models, deep learning}
\maketitle
\section{Introduction}\label{sec1}
In actuarial literature for non-life pricing deep learning models (see e.g.~\cite{goodfellow_deep_2016}) 
have gained great popularity in recent years. 
But still, generalized linear models (GLM) are the current standard regression and classification models 
for insurance problems in many companies. 
The reasons for this are manifold and range from interpretability 
to the well-established data preparation pipelines with the corresponding data revision processes 
as well as a lot of experience regarding the model update process. 
Restructuring these entrenched processes not only takes a lot of time but also requires 
a significant amount of effort in terms of documentation for the authorities.
In this manuscript, we try to bridge the gap between the current industry standard models 
and the current state-of-the-art transformer model by~\cite{vaswani_attention_2017}, 
which is currently the industry standard model for image, audio, and text processing.
To make a smooth transition period possible for the industry, we not only use the 
predictive power of the transformer model as an argument for the change but also
blend the current models and typical data pipeline characteristics into these transformer models. 
Either to keep a familiar interpretation 
or the same data preparation as currently in place at the companies. \\
There are typically three types of problems in non-life insurance pricing: 
claims frequency prediction (usually done by Poisson GLMs), the prediction of claim severity (typically Gama GLMs), and logistic regression are used for all types of demand models like churn and conversion modeling.
Here, we create a new model structure that benefits from the well-established GLM processes 
currently in place for insurance companies and use the great predictive power of the now 
well-established transformer model. 
We compare the results of prevalent real-world example data, with the industry standard models (GLMs) and the current 
present models in research for insurance when it comes to neural networks.\\
\subsection{Organization of this paper}
Since our models are built as a combination of existing models 
we first give a brief overview of the related work in the next section.
The following chapter describes the different model approaches.
After that, we present the results of our experiments on a real-world claim frequency dataset 
as well as go through the implementation details. Finally, we conclude 
with a discussion of the results and an outlook on future work.
\section{Related work}\label{sec2}
Since they are still the predominant tool in pricing actuarial departments, we start with the 
backbone of the industry, the GLM, 
which was introduced by~\cite{nelder_generalized_1972} 
and further elaborated upon in the book~\cite{mccullagh_generalized_1983}. 
Since it is also common practice to enhance these models by splines and regularization, we also refer 
here to the generalized additive model (GAM) 
(see~\cite{hastie_generalized_1986} and~\cite{hastie_generalized_1990}). 
For a variety of methods in the actuarial field, 
we highly recommend the following book by~\cite{wuthrich_statistical_2023}, 
which will also be used as the basis for a lot of benchmark models described in this manuscript. 
In the literature, there are several methods published that try to combine the GLM methods 
and neural networks. Examples are the models DeepGLM and the 
DeepGLMM of~\cite{tran_bayesian_2018}, which are transforming the inputs via a neural network 
before building a GLM or a GLMM on top. In actuarial literature, we want to mention here the 
Combined Actuarial Neural Network (CANN) as such an approach, see here e.g.~\cite{schelldorfer_nesting_2019}. 
There, the authors describe how to nest existing GLMs into feed-forward neural networks (FFN) 
with skip connections, either by setting the GLM $\beta$'s as trainable weights or as not trainable 
weights and optionally using credibility weights. 
Furthermore, in the actuarial field recently, the LocalGLMnet was introduced by~\cite{richman_localglmnet_2023}, 
which preserves the linear structure of the GLM but uses feature-dependent linear predictors $\beta(\mathbf{x})$, 
which are the result of a multilayer perceptron (MLP). The naming convention comes here from 
the assumption that in a small environment around $\mathbf{x}$ the $\beta(\mathbf{x})$ can be approximated by a constant $\beta$, 
and, therefore, can be interpreted as a local {GLM}. A very similar architecture would be a simplified ResNet~\cite{he_deep_2016} 
as it is described, {e.g.} by~\cite{gorishniy_revisiting_2021}.
In regards to pure neural networks for supervised learning on tabular data 
there has been a lot of progress made in the previous years. 
A pivotal role in this evolution has been played here the TabNet by~\cite{arik_tabnet_2020}, 
which uses sequential attention to reweigh features during the training process. 
It's worth noting that the methods employed prior to 2021 have been 
comprehensively summarized in the work of~\cite{gorishniy_revisiting_2021}.
A very powerful approach for small datasets is the Prior-Data Fitted Network TabPFN introduced by~\cite{hollmann_tabpfn_2023}
that (once pre-trained) does not need training but instead uses the whole training and test 
samples as inputs and yields predictions for the entire test set in a single forward pass. 
The main approach that our work is based on is a transformer-based model~\cite{vaswani_attention_2017} 
that is adapted to fit tabular data structures called feature tokenizer transformer (FTT), which was introduced 
by~\cite{gorishniy_revisiting_2021}. These models were enhanced in~\cite{gorishniy_embeddings_2022}, 
where it was demonstrated that by improving the embeddings, a substantial impact on the performance could be attained. 
\section{Model approaches}\label{sec_model_architecture}
In this section, we introduce the current leading models in the industry 
as well as our adaptions of the transformer models. 
Like in the case of the LocalGLMnet, we also try to reuse well-established Models as much as possible. 
We therefore anticipate that our approaches will lead to models 
that are conceptually intuitive and require less exertion to attain excellent performance.
Implementation details for all the models are provided in section~\ref{sec_implementation}.
For our supervised learning problems, we consider here the usual notation for Datasets 
$D = {\{\left(y_{i}, \mathbf{x}_{i}, v_{i}\right)\}}_{i=1}^{n}$ where for the $i$'th observation, $v_i>0$ represents the given exposure, 
$y_i$ is the response and the covariates $\mathbf{x}_i$ 
are represented by a vector of $k$ numerical as well as categorical features. 
\subsection{GLMs and GAMs}\label{subsec_glm_gam}
We use here the notation of~\cite{wuthrich_statistical_2023}. We assume here that all covariables $\mathbf{x}_i$ 
are numerical (i.e. $\in \mathbb{R}^k$), so for example, a dummy or one hot encoding is applied beforehand. 
Then, the prediction via a GLM would be given 
\[\hat{y} = g^{-1}(\beta_0 +  \langle\beta,\mathbf{x}\rangle)\]
where $\langle.,.\rangle$ describes a scalar product, the $\beta_0$ is the bias (also called intercept), 
$\beta \in \mathbb{R}^k$ is the regression parameter vector, where for every feature $x_{j}$ with $j \in \{1,\ldots,k\}$ 
we have a corresponding regression parameter 
$\beta_j\in \mathbb{R}$ and $g()$ is called the link function. 
The main assumption for a glm is that $Y$ follows an exponential dispersion family (EDF). 
In the case of a Poisson regression with exposure $v_i$, for $i \in \{1,\ldots,n\}$ 
the $i$'th prediction would be given by:
\[\hat{y}_i = \exp(\beta_0 +  \langle\beta,\mathbf{x}_i\rangle)v_i\]
Note that $g()$ is usually chosen for theoretical reasons as the canonical link function 
or it is chosen differently for practical implementation reasons. 
A typical example of such a deviation in the actuarial world would be a log link for a Gamma GLM despite 
it not being the canonical link, because one prefers the multiplicative nature.
Of course, if one deviates from the canonical link, one has to be aware of the missing balance property 
(see section 5.1.5 in~\cite{wuthrich_statistical_2023}). 
Note that it is very typical to include interactions into GLMs as well as to enhance
the GLM linear predictor $\eta = \langle\beta, \mathbf{x}\rangle$ 
by splines $s(x_j)$ for $j \in {1,\ldots,k}$. We would argue that the actual industry-standard are currently 
generalized additive models (GAMs) where the prediction would be given by 
\[ \hat{y} = g^{-1}\left(\beta_0 +  \sum_{j=0}^{k} \beta_j s(x_{j})\right).\] 
One further important point to notice is that data preprocessing 
is clearly one of the most crucial parts of current actuarial modeling. 
{E.g.} it is very common to convert numerical features into categorical ones via binning 
and grouping original categorical feature values. 
The GLMs and GAMs obviously have the big advantage of being a very transparent model type when it comes to 
interpretability because the feature contribution of the linear predictor 
for $j$'th feature and $i$'th observation is directly given by $\beta x_{ij}$.
\subsection{Feed-forward neural network}\label{subsec_fnn}
As usual, we refer here to~\cite{goodfellow_deep_2016} for more insights into Feed-forward neural networks (FNN). 
Due to the dominance of the GLMs in non-life insurance, we define here the FNN 
in a similar way as as done in~\cite{wuthrich_generalized_2019} and in the Book~\cite{wuthrich_statistical_2023}, 
where the FNN is seen as a generalization of a {GLM}. 
Here, given a depth $d \in \mathbb{N}$ and the last activation function $h$ the predictor of a FNN is given by: 
\[\hat{y} = h\left(w^{(d+1)}_0 +  \left\langle w^{(d+1)},\left(z^{(d)}\circ\ldots\circ z^{(1)}  \right)(\mathbf{x})\right\rangle\right)\] 
where for $m \in \{1,\ldots,d\}$, the $m$'th Layer is defined as a function:  
\[z^{(m)}: \mathbb{R}^{q_{m-1}} \rightarrow \mathbb{R}^{q_{m}}\] 
\begin{equation}\label{eq:fnn_dense_layer}
    x \mapsto z^{(m)}(x) = {\left(z_{1}^{(m)}(x),\ldots,z_{q_m}^{(m)}(\mathbf{x})\right)}^{T}, 
\end{equation}
with activation function $\phi_m: \mathbb{R} \rightarrow \mathbb{R}$, weights $w_j^{(m)} \in \mathbb{R}^{q_{m-1}}$ 
and bias $w_{0,j}^{(m)} \in \mathbb{R}$ the prediction of the neurons of the $m$'th Layer $z_{j}^{(m)}(\mathbf{x})$ are defined as: 
\[z_{j}^{(m)}(\mathbf{x}) = \phi_m\left(w_{0,j}^{(m)}+\left\langle w_j^{(m)},\mathbf{x}\right\rangle \right). \]
In the context of a Poisson regression with exposure $v_i$, for $i \in \{1,\ldots,n\}$ 
the $i$'th neural network prediction would be given by:
\[\hat{y}_i = \exp\left(w^{(d+1)}_0 +  \left\langle w^{(d+1)},\left(z^{(d)}\circ\ldots\circ z^{(1)}  \right)(\mathbf{x}_i)\right\rangle\right)v_i\] 
with $w^{(d+1)}_0 \in \mathbb{R}$ and $w^{(d+1)} \in \mathbb{R}^{q_d}$. 
The FNNs are fitted via variations of stochastic gradient descent (SGD), and usually, to prevent overfitting, one applies an 
early stopping criteria, see here for example~\cite{goodfellow_deep_2016}. 
\subsection{Combined actuarial neural network approach}\label{subsec_cann}
The Combined actuarial neural network (CANN) approach was introduced in~\cite{wuthrich_editorial_2019} 
and details are given for example in tutorials~\cite{schelldorfer_nesting_2019} and~\cite{wuthrich_generalized_2019} 
as well as the book~\cite{wuthrich_statistical_2023}.
Here, one defines the prediction as: 
\begin{equation}\label{eq:cann_pred}
    \hat{y} = g^{-1}\left(\beta_0 +  \langle\beta,\mathbf{x}\rangle + w^{(d+1)}_0 +\left\langle w^{(d+1)},\left(z^{(d)}\circ \ldots \circ z^{(1)}  \right)(\mathbf{x})\right\rangle\right)
\end{equation}
The first term $(\beta_0 +  \langle\beta,\mathbf{x}\rangle)$ represents the linear predictor of a GLM and the remaining term 
on the right-hand side describes the output of the FNN (before applying the last activation function).
This model can be seen as an extension of an already-fitted GLM.\ Given already fitted GLM parameters $\beta_0$ 
and $\beta$, one initializes the model by those GLM parameters, and for the weights of the last layer of the FNN,  
starts with a zero parametrization. This way, one starts with the same prediction as the GLM and enhances the GLM 
prediction further by the nonlinear features representation of the FNN and more complex interactions. 
Note that usually, the betas of the GLM are here set as non-trainable. In such a case, one can interpret this model 
as a kind of residual model. In the case of Poisson regression with exposure $v_i$, for $i \in \{1,\ldots,n\}$ 
one can also fit the FNN with the last layer exponential activation:  
\begin{equation}\label{eq:cann_poisson_pred}
    \hat{y} = \exp\left(w^{(d+1)}_0 +\left\langle w^{(d+1)},\left(z^{(d)}\circ\ldots\circ z^{(1)}  \right)(\mathbf{x})\right\rangle\right)\cdot v^{GLM}
\end{equation}
with $v^{GLM}$ defined as the exposure times the GLM prediction
\begin{equation}\label{eq:exposure_x_glm}
    v^{GLM}:= v_i \cdot \exp(\beta_0 +  \langle\beta,\mathbf{x}\rangle).
\end{equation}
\subsection{LocalGLMnet}\label{subsec_localGlMnet}
The LocalGLMnet was introduced by~\cite{richman_localglmnet_2023} and tries to enhance the GLM 
by using coefficients that are dependent on the covariables $\beta(x)$ and determined by a {FNN}. 
So, the LocalGLMnet prediction can be defined as follows:
\begin{equation}\label{eq:localglmnet_pred} 
    \hat{y} = g^{-1}(\beta_0 +  \langle\beta(x),\mathbf{x}\rangle),
\end{equation}
where $\beta(x)$ are the result of a FNN of depth $d$ and same output and input dimension: 
\[\beta: \mathbb{R}^{k} \rightarrow \mathbb{R}^{k}\] 
\begin{equation}\label{eq:localGlMnet_betas}
    \beta(\mathbf{x}) = \left(z^{(d)}\circ\ldots\circ z^{(1)}  \right)(\mathbf{x}).
\end{equation}
There are many ways to interpret this model, e.g.\ a special kind of 
varying coefficient model~\citep{hastie_varying-coefficient_1993}. The main idea for the naming notation 
is that in a small environment around $\mathbf{x}$ the $\beta(\mathbf{x})$ can be approximated by a constants $\beta$,
and therefore can be interpreted as a local GLM.\ Note that the approach of using a skip connection is present also 
in many other models like, for example, the ResNet~\citep{he_deep_2016}.
If a GLM model with regression parameters $\beta^{GLM}_0$ and $\beta^{GLM}$ is already in place 
we can initialize the weights of the FNN such that we start with the GLM prediction. 
As described in great detail in~\cite{richman_localglmnet_2023} one can later visualize the model attentions $\beta_{j}(\mathbf{x})$ 
and model contribution $\beta_j(\mathbf{x})x_j$ of a feature $j\in\{1,\ldots,k\}$ and get this way an inside into the behavior of the model. 
Which makes this model interpretable on an observation level 
and therefore very attractive in the actuarial field. 
\subsection{Feature Tokenizer Transformer}\label{subsec_ftt}
The Feature Tokenizer Transformer (FTT) is a transformer-based model~\citep{vaswani_attention_2017} 
but adapted to fit tabular data structures. It was introduced by~\cite{gorishniy_revisiting_2021}.
The first part of the transformer is a so-called Feature Tokenizer. 
This Feature Tokenizer converts each covariate input (categorical and numerical) 
into vectors of fixed length called embedding dimension $d_{emb}$. 
Let $x \in \mathbb{X}$ be a covariate vector with $k$ features $x = {(x_1,\ldots,x_k)}^T$ 
where $x_j \in \mathbb{X}_j$ is the $j$'th feature with $j \in \{1,\ldots,k\}$. 
Then, the Feature Tokenizer for the $j$'th feature is defined as a function 
\[FT_j(x_j): \mathbb{X}_j \rightarrow \mathbb{R}^{1\times d_{emb}}\]
The embedding for each feature consists of trainable weight matrices $W_j$ and biases $b_j$. 
Let $k_{num}$ and $k_{cat}$ be the number of numerical and categorical features and let w.l.o.g.\ 
$j \in \{1,\ldots,k_{cat}\}$ be the categorical features and $j \in \{k_{cat}+1,\ldots,k\}$ be the numerical ones. 
In the categorical case, the implementation is done by a lookup table for the weights. 
So formally for a categorical feature $x_j \in \mathbb{X}_j$ with $j \in \{1,\ldots,k_{cat}\}$ 
and number of unique feature values $S_j=|\mathbb{X}_j|$ the embedding is given by:
\[FT_j(x_j) =  e_j^{T} W_j + b_j,\]
where $e_j$ is the one hot vector, $W_j \in \mathbb{R}^{S_j\times d_{emb}}$ and $b_j \in \mathbb{R}^{1\times d_{emb}}$.
Different from other tabular transformer techniques such as e.g.~the TabTransformer~\citep{huang_tabtransformer_2020} 
the FTT does not only embed the categorical features but also the numerical ones. 
In the numerical case, the embedding is done by a FNN with one linear output layer of dimension $d_{emb}$. 
So for a numerical feature $x_j \in \mathbb{R}$ with $j \in \{k_{cat}+1,\ldots,k\}$ the embedding is given by:
\[FT_j(x_j) =  x_j W_j + b_j,\]
where $x_j \in \mathbb{R}$, $W_j \in \mathbb{R}^{1\times d_{emb}}$ and $b_j \in \mathbb{R}^{1\times d_{emb}}$.
At this point, it's important to note that as~\cite{gorishniy_embeddings_2022} 
demonstrated, there are several other ways to implement numerical embedding. 
For example, by using different piecewise linear encoding techniques 
such as bins from quantiles or target-aware bins as well as techniques using periodic activation functions. 
In fact, it is very popular in insurance to use bins for numerical features and add numerical features this way as categorical ones into a 
GAM framework due to technical software restrictions by common insurance pricing software. 
So, if such a binning is already in place, one can also simply use a categorical embedding as described above.
The authors showed in~\cite{gorishniy_embeddings_2022} that a good chosen numerical embedding 
can have a substantial benefit on the performance of the model.
After the feature tokenization, those embedded features-tensors 
are then stacked upon each other in addition to a random initialized trainable weight called $cls$ token with $cls \in \mathbb{R}^{1\times d_{emb}}$. 
\[Stack_{Input}(\mathbf{x}) := \begin{bmatrix}
    cls \\
    FT_1(x_1) \\
    \vdots \\
    FT_k(x_k)
    \end{bmatrix} \in \mathbb{R}^{(k+1)\times d_{emb}}\]
This stacked input is then passed through a sequential array of $t$ transformer-encoder blocks.
\[Stack_{Output}(\mathbf{x}):= \left(T^{Block}_t\circ\ldots\circ T^{Block}_1  \right)(Stack_{Input}(\mathbf{x})),\]
where a transformer block $T^{Block}_l$ for $l \in \{1,\ldots,t\}$ 
\[T^{Block}_l(x): \mathbb{R}^{(k+1)\times d_{emb}} \rightarrow \mathbb{R}^{(k+1)\times d_{emb}}\] 
is defined as a combination of a multi-head attention layer and a feed-forward network (FFN) layer
as well as additional layer normalizations and residual additions in between. 
For a description of the transformer-encoder block, we refer to~\cite{vaswani_attention_2017}. 
For the prediction, the row with the same position as the cls token in $Stack_{Input}(x)$ (usually first or last, here w.l.o.g.\ first) 
of the output matrix $Stack_{Output}(\mathbf{x}) \in \mathbb{R}^{(k+1)\times d_{emb}}$ is taken ${Stack_{Output}(\mathbf{x})}_1 \in \mathbb{R}^{1 \times d_{emb}}$
and then passed through a layer normalization, an elementwise ReLU activation, and a dense last layer with an activation function $h$ of choice.
\[\hat{y} = h\left(w_0 +  \left\langle w, ReLU{\left(Norm_{Layer}({Stack_{Output}(\mathbf{x})}_1)\right)}^T \right\rangle\right) \]
Within the framework of a Poisson regression with 
exposure denoted as $v_i$ for $i$ ranging from $1$ to $n$, 
the prediction of the $i$'th observation can be expressed as follows:
\begin{equation}\label{eq:ftt_poisson_pred} 
    \hat{y}_i = \exp\left(w_0 +  \left\langle w, ReLU{\left(Norm_{Layer}({Stack_{Output}(\mathbf{x}_i)}_1)\right)}^T \right\rangle\right)v_i,
\end{equation}
with $w_0 \in \mathbb{R}$ and $w \in \mathbb{R}^{d_{emb}}$.
For more details on the exact implementation 
of the FTT and hyperparameter choice used in this paper, we refer to section\ \ref{sec_experiments}. 
\subsection{Combined actuarial feature tokenizer transformer}\label{subsec_cann_ftt}
The Combined actuarial feature tokenizer transformer (CAFTT) is defined in the same way as the CANN approach 
in section\ \ref{subsec_cann} but instead of using a FNN 
we are using here a FTT on the right-hand side of equation~\ref{eq:cann_pred}.
So, the prediction can be described as follows: 
\begin{equation}\label{eq:cann_ftt_pred}
    \hat{y} = g^{-1}\left(\beta_0 +  \langle\beta,\mathbf{x}\rangle + w_0 +  \left\langle w, ReLU{\left(Norm_{Layer}({Stack_{Output}(\mathbf{x})}_1)\right)}^T \right\rangle\right)
\end{equation}
As in the case of the CANN model, this model can be seen as an extension of an already fitted {GLM}.
Given already fitted GLM parameters $\beta_0$ and $\beta$, 
one also initializes here the weights of the prediction layer ($w_0$ and $w$) with a zero parametrization and starts this way 
the training with the same prediction as the {GLM}. 
During training, one enhances the GLM prediction further by additional nonlinear feature representation 
and interactions built via the transformer. 
Note as before in section~\ref{subsec_cann}, when using here non trainable GLM betas, 
one can interpret this model as a kind of residual model. {E.g.} 
in the Poisson regression case with exposure $v_i$, for $i \in \{1,\ldots,n\}$ one can also 
fit here the FTT with exponential activation at the last layer:  
\begin{equation}\label{eq:cann_ftt_pred_poisson}
    \hat{y} = \exp\left(w_0 +  \left\langle w, ReLU{\left(Norm_{Layer}({Stack_{Output}(\mathbf{x})}_1)\right)}^T \right\rangle\right)\cdot v^{GLM}
\end{equation}
with $v^{GLM}$ defined as in equation~\ref{eq:exposure_x_glm}. 
\subsection{LocalGLM-Feature-Tokenizer-Transformer}\label{subsec_localGlMnet_ftt}
The LocalGLM-Feature-Tokenizer-Transformer (LocalGLMftt) is a combination of the LocalGLMnet approach by~\cite{richman_localglmnet_2023} 
(see section~\ref{subsec_localGlMnet}) and the FTT by~\cite{gorishniy_revisiting_2021} (see section\ \ref{subsec_ftt}).
So we start with the same prediction formula as in equation\ \ref{eq:localglmnet_pred} 
\begin{equation}\label{eq:localglmnet_ftt_pred} 
    \hat{y} = g^{-1}(\beta_0 +  \langle\beta_{FTT}(x),x\rangle)
\end{equation}
but the $\beta_{FTT}(x)$ are now the result of a FTT with $k$ dimensional output.
\begin{equation}\label{eq:localglmnet_ftt_betas} 
    \beta_{FTT}(x): = z\left(ReLU{\left(Norm_{Layer}({Stack_{Output}(x)}_1)\right)}^T \right)
\end{equation}
with $z: \mathbb{R}^{d_{emb}}\rightarrow \mathbb{R}^{k}$ being a dense layer (see equation~\ref{eq:fnn_dense_layer}) with output dimension $k$ and linear action function. 
This way, we can use the same interpretation as well as visualization techniques as in the LocalGLMnet model case.
\section{Experiments on real world data}\label{sec_experiments}
In this section, we compare the different models on a real-world dataset. Here we describe the dataset and the data preprocessing, 
go through the model implementation details, and finally present the results.
\subsection{Dataset and data preprocessing:}\label{sec_dataset}
The dataset of choice is the well-known French motor third-party liability (MTPL) 
claims frequency dataset FreMTPL2freq~\cite{dutang_casdatasets_2018}. 
It is used as a standard benchmark for a lot of model applications in the actuarial field. 
It is used by~\cite{lorentzen_peeking_2020} in the context of interpretability, 
regarding the nesting of GLMs into the FNNs the above mentioned~\cite{wuthrich_generalized_2019},~\cite{richman_localglmnet_2023}, 
and the book~\cite{wuthrich_statistical_2023} all use this dataset. 
It is important to note that we are using here \texttt{version 1.0-8} of the dataset, 
as it is also used in the mentioned literature. \\
\textbf{Data-Exploration}: The data (before data cleaning) consist of $678,013$ observations 
with an ID, response $y$, an exposure column $v$ as well as
9 covariate columns $\mathbf{x}$, two of which are categorical, 6 numerical, and one bool.
Due to the very common use in non-life pricing literature, we forego here on a detailed description of the dataset
and refer to the tutorial by~\cite{noll_case_2020}.\\
\textbf{Data-Cleaning}: 
Furthermore, we are using the same data cleaning process as described in Appendix B~\cite{wuthrich_statistical_2023}. \\
\textbf{Data-Preperation}: 
Regarding the data preparation we follow for the Benchmark GLM models the same feature engineering 
as sections 5.2.4 and 5.3.4 of~\cite{wuthrich_statistical_2023}. 
For the benchmark FNN and LocalGLMnet, we use the same preprocessing as described in~\cite{richman_localglmnet_2023} 
by applying a standard scaler to the numerical features and a one-hot encoding to the categorical features. 
For the transformer-based models, we also use the same numerical feature preprocessing as in the LocalGLMnet case 
but since we are applying a categorical embedding, we don't need a one hot encoding for the categorical features.\\
\textbf{Train/Test-Split}: The majority of the manuscripts using this dataset are using the same train/test split,
which is described in Listing 5.2 of the book~\cite{wuthrich_statistical_2023}. 
The test set contains $67,801$ observations and is only used for the final hold-out test. 
Whereas the train set contains $610,206$ observations and is used either completely for the model fitting ({e.g.}~in, the case of the GLMs)
or in the case of machine learning models, again split into 90\% for model training and 10\% (validation-set) 
for hyperparameter-fitting as well as early stopping.\\
\textbf{Reproducibility}: 
To quickly reproduce our results, we provide besides the Python code for the model approaches
also the detailed code for the data preprocessing (see section~\ref{sec_implementation}).\\
\subsection{Model implementation details}\label{sec_implementation}
\textbf{Hardware and software}:
The Hardware used was a \texttt{Intel(R) Xeon(R) CPU @ 2.20GHz} CPU with \texttt{13GB} of System-RAM, 
and a \texttt{NVIDIA Tesla T4} GPU with \texttt{15GB} of {GPU-RAM}.
We implemented the models in \texttt{Python}~\cite{python} (\texttt{version: 3.10.12}) 
using mainly the \texttt{scikit-learn} API~\cite{sklearn_api} (\texttt{version: 1.2.2}) for the GLMs and the 
deep learning framework \texttt{tensorflow}~\cite{tensorflow2015-whitepaper} (\texttt{version: 2.14.0}). 
Further details regarding the exact implementation and used environment can be found in the provided code available at \href{https://github.com/BrauerAlexej/}{https://github.com/BrauerAlexej/}\\
\textbf{Loss function and metrics}:
We are using here the Poisson deviance loss as the loss function for fitting and loss metric for evaluation: \\
\begin{equation}\label{eq:poisson_deviance_loss}
    \mathcal{L}_{dev}(\hat{y},y) := \sum_{i=1}^{n} 2\left(y_i \log \frac{y_i}{\hat{y}} - y + \hat{y}\right),
\end{equation}
where $y$ is the observed claims frequency and $\hat{y}$ is the 
predicted claims frequency already multiplied by the exposure $v_i$.
For insights into the properties of this loss function under the Poisson model, 
we refer to Section 4.1.3 in the book~\cite{wuthrich_statistical_2023}.
\\
\textbf{GLM}:
Because we are dealing with the prediction of claims frequency,
we train Poisson GLMs with a log link function.
As an industry benchmark, we fitted the three GLMs described 
in section 5.3.4 of the book by~\cite{wuthrich_statistical_2023}.
The difference between the 3 GLMs is the feature engineering.
The first GLM ($GLM_1$) uses all features after the initial feature preparation and dummy encoding. 
The second GLM ($GLM_2$) is based on the first one, 
but where GLM1 uses a binning for a certain numerical feature, 
GLM2 uses more complex functional forms for this feature 
(involving logarithmic and polynomial transformations).
The third GLM ($GLM_3$) is based on the second one 
but includes interaction terms between two features.
The unpenalized GLMs are fitted with the \texttt{scikit-learn} API~\citep{sklearn_api} 
using the Newton-Cholesky solver. 
\\
\textbf{FNN}: 
Because we are dealing with the prediction of claims frequency,
we train two Poisson neural networks as described in chapter~\ref{sec_model_architecture}.
The first FNN ($FNN_{OHE}$) is fitted on one-hot encoded feature inputs. 
The dimension of the input feature is here $k=40$.
For the second FNN ($FNN_{EMB}$) we include the categorical features 
via two-dimensional embeddings instead of one-hot encoding them. 
We chose for both models the same backbone architecture 
as described in section 3.3.2 of~\cite{wuthrich_generalized_2019}, namely 
besides the input layers, 3 hidden layers dimension $(q_{1}, q_{2}, q_{3}) = (20, 15, 10)$ 
except for the last one, we chose every activation function as a hyperbolic tangent function and the output 
layer as an exponential activation function.
For the initialization, we set the weights of the last hidden layer to 
zero and the bias to the log of the mean response on the training dataset. The batch size was set to $5000$ in the case of the $FNN_{OHE}$ and to $7000$ for the $FNN_{EMB}$ model. 
In both cases, we fit the model on 500 epochs using the \texttt{nadam} 
optimizer and an early stopping criteria with a patience of 15 on the 10\% validation data.
\\
\textbf{CANN}: 
For the poisson implementation of the $CANN$ model (see equation~\ref{eq:cann_poisson_pred}), 
we used as the base FNN the same structure as $FNN_{EMB}$ and set 
the $v^{GLM}$ (see equation~\ref{eq:exposure_x_glm}) via the $GLM_3$. 
For the initialization, we set the weights and the bias of the last hidden layer to zero. 
This way, we start with the same prediction as $GLM_3$.
Regarding the fit, we chose here the same optimizer and early stopping 
criteria hyperparameter as in the $FNN_{EMB}$ case.
\\
\textbf{LocalGLMnet}:
For the $LocalGLMnet$ model, we implement a new GLM that 
uses instead of dummy encoding a one-hot encoding for the categorical features, 
see here regarding this choice section 3.6 in~\cite{richman_localglmnet_2023}.
As one would expect, this model performs very similarly to $GLM_1$. 
This start GLM is then used as the base of the $LocalGLMnet$. 
Regarding the FNN for the $\beta(\mathbf{x})$ (see equation~\ref{eq:localGlMnet_betas}), 
we chose the same hidden architecture as in the $FNN_{OHE}$, but we used here as the output layer dimension $k=40$.
For the initialization, we set the weights of the last layer in $\beta(\mathbf{x})$ to zero 
and the bias to those of the start {GLM}. 
The trainable weight $\beta_0$ of equation~\ref{eq:localglmnet_pred} is initialized by the intercept of the start {GLM}. 
For the training, we chose here the same optimizer and early stopping criteria hyperparameter as in the $FNN_{OHE}$ case.
\\
\textbf{FTT}:
For the FTT, we used mostly the default hyperparameters described 
in the appendix~\cite{gorishniy_revisiting_2021}.
Namely, we used 3 transformer blocks with $h=8$ attention heads. 
As usually done, we chose for the key and value dimension 
of the multi-head attention layer $d_{value} = d_{key} = d_{emb}/h$~\citep{vaswani_attention_2017}. 
The activation function of the FFN inside the transformer is set to ReGLU~\citep{shazeer_glu_2020}. 
For the dimension of the hidden layer, we also chose $d_{ffn} = 4/3 \cdot d_{emb}$, 
but since the ReGLU function splits the dimension of the input tensor in two, 
the dimension of the first hidden layer is set to $2\cdot d_{ffn}$.
The dropout rate for the attention layer is set to $0.2$ and for the FFN layer to $0.1$.
A difference to the~\cite{vaswani_attention_2017} transformer is that
they used a prenormalization setting instead of a postnormalization.
However, for performance reasons, the pre-normalization layer is not implemented for the first transformer block.
As shown in equation~\ref{eq:ftt_poisson_pred}, we used for the output layer the exponential activation function.
Regarding the embedding dimension, we deviated from the default settings in~\cite{gorishniy_revisiting_2021}. 
To reduce training time, we used a smaller embedding dimension of $d_{emb}=32$ instead of $d_{emb}=192$.
The weights and biases are initialized with a uniform distribution in the range $[-1/\sqrt{d_{emb}},1/\sqrt{d_{emb}}]$. 
\\
The reason for fitting this model on the default hyperparameters is that
in~\cite{gorishniy_revisiting_2021} the authors showed in Table 4, 
that the performance on 10 different datasets of the hyperparameter tuned $FTT_{tuned}$
is very comparable to an ensemble of 5 $FTT_{def}$ models with default hyperparameter settings. 
Although hyperparameter tuning is these days due to packages like \texttt{Optuna}~\citep{optuna_2019} 
very accessible, it is still for transformer models very time-consuming. 
So, we decided to accept the small gain loss and perform no hyperparameter tuning.
In the next chapter, one can see the average performance 
of the single $FTT_{def}$ (with default setting) vs. an ensemble of 5 $FTT_{def}$ models.
\\
We fit the models on 500 epochs with a batch size of 1024, using the \texttt{AdamW} optimizer 
with learning rate $1\mathrm{e}\text{-}4$, weight decay $1\mathrm{e}\text{-}5$ and 
early stopping criteria with a patience of 15 on the 10\% validation data. 
We have provided our code via GitHub, which is extensively commented on to provide detailed explanations of the steps and subclasses. The code is implemented using the subclassed \texttt{tensorflow} framework.
\\
\textbf{CAFTT}:
The CAFTT model is implemented in the same way as the $FTT_{def}$ model 
but as in the CANN model case, we set the $v^{GLM}$ (see equation~\ref{eq:exposure_x_glm}) 
via the $GLM_3$ and initialized the weights $w$ and bias $w_0$ 
in equation~\ref{eq:cann_ftt_pred_poisson} to zero. 
\\
\textbf{LocalGLMftt}:
We start with the same start GLM model as in the case of the $LocalGLMnet$. 
For the $\beta_{FTT}(\mathbf{x})$ (see equation~\ref{eq:localglmnet_ftt_betas}) 
a $FTT_{def}$ is fitted with the same output dimension as features $k=9$.
The last layer weights of this FTT are set to zero, and the bias is set to one. 
Note that for a categorical feature $j$ the value of $\beta_{FTT}(\mathbf{x})_j$ is in $\mathbb{R}$, 
but the start GLM uses the one hot encoding. 
We calculate there for the scalar product of equation~\ref{eq:localglmnet_ftt_pred} 
and the initialization with the start GLM betas via applying categorical embeddings. 
For the training loop, we chose here the same optimizer and 
early stopping criteria as in the $FTT_{def}$ case.
\subsection{Results}\label{sec_results}
As mentioned above, we fitted Poisson GLMs and Poisson neural networks 
with a log link function or exponential activation function at the end. 
For every model type, we fitted 15 models with different seeds. 
In the case of models involving neural networks, 
we used for every of the 15 models an own validation split for early stopping.
The holdout test set was only used for the final evaluation and 
was the same for each model (see chapter~\ref{sec_dataset}). 
In table~\ref{tab:model_fitting_properties}, we present the properties of model fits.
For every model, the average and standard deviation 
regarding the run-time on the hardware used,
the number of trainable parameters/weights 
and for the neural network-based models, the epoch of the best model 
on the validation loss according to early stopping is displayed. 
\begin{table}[h]
    \caption{Model fitting properties}\label{tab:model_fitting_properties}
    \begin{tabular*}{\textwidth}{@{\extracolsep\fill}lrlr}
    \toprule%
    model                   & epochs            & run-time in $mm\text{:}ss.ms$        & \#parameters \\
    \midrule
    (a) mean model & - & $00\text{:}00.05\;(\pm00\text{:}00.00)$ & $1$ \\
    (b) $GLM_1$ & - & $00\text{:}02.22\;(\pm00\text{:}00.47)$ & $49$ \\
    (c) $GLM_2$ & - & $00\text{:}02.75\;(\pm00\text{:}00.97)$ & $48$ \\
    (d) $GLM_3$ & - & $00\text{:}01.90\;(\pm00\text{:}00.41)$ & $50$ \\
    (e) $FFN_{OHE}$ & $42\,(\pm15)$ & $00\text{:}37.81\;(\pm00\text{:}08.62)$ & $1,306$ \\
    (f) $FNN_{EMB}$ & $73\,(\pm22)$ & $00\text{:}58.73\;(\pm00\text{:}13.91)$ & $792$ \\
    (g) $CANN$ & $90\,(\pm54)$ & $01\text{:}08.56\;(\pm00\text{:}33.16)$ & $792$ \\
    (h) $LocalGLMnet$ & $25\,(\pm08)$ & $00\text{:}29.72\;(\pm00\text{:}05.10)$ & $1,737$ \\
    (i) $FTT_{def}$ & $79\,(\pm17)$ & $26\text{:}09.86\;(\pm05\text{:}02.32)$ & $27,133$ \\
    (j) $CAFTT_{def}$ & $57\,(\pm14)$ & $19\text{:}30.16\;(\pm03\text{:}40.45)$ & $27,133$ \\
    (k) $LocalGLMftt_{def}$ & $53\,(\pm16)$ & $19\text{:}47.01\;(\pm05\text{:}10.65)$ & $27,430$ \\
    \botrule
    \end{tabular*}
    \footnotetext{Average and standard deviation of model fitting properties  
    over 15 fits with different seeds.}
    \end{table}
\noindent
As one would expect, the average run-time for the GLMs, which is around 2s,
is very low compared to the neural network-based models, 
where the runtimes range between 30s to 69s for the FFN-based models and from 
19min to 26min for the transformer-based implementations. 
This is due to the immense amount of parameters that have to be fitted 
in the case of the neural network-based models, 
which is extremely prevalent in the case of the transformer-based models, 
where the number of trainable parameters is around 27k,
see here for the right column of table~\ref{tab:model_fitting_properties}.
\\
\\
In table~\ref{tab:model_results}, we report the average Poisson deviance loss as well as the standard deviation 
on the train and test dataset. 
\begin{table}[h]
    \caption{Average single model results}\label{tab:model_results}
    \begin{tabular*}{\textwidth}{@{\extracolsep\fill}llll}
    \toprule%
    model                   & Train-loss in \%   & Test-loss in \%           & $avg(\hat{y})$ in \% \\
    \midrule
    (a) mean model & $25.213$ & $25.445$ & $7.363$ \\
    (b) $GLM_1$ & $24.101$ & $24.146$ & $7.390$ \\
    (c) $GLM_2$ & $24.091$ & $24.113$ & $7.398$ \\
    (d) $GLM_3$ & $24.084$ & $24.102$ & $7.405$ \\
    (e) $FFN_{OHE}$ & $23.754\,(\pm0.033)$ & $23.865\;(\pm0.016)$ & $7.431\;(\pm0.121)$ \\
    (f) $FNN_{EMB}$ & $23.768\,(\pm0.016)$ & $23.827\;(\pm0.015)$ & $7.424\;(\pm0.109)$ \\
    (g) $CANN$ & $23.742\,(\pm0.061)$ & $23.810\;(\pm0.033)$ & $7.444\;(\pm0.110)$ \\
    (h) $LocalGLMnet$ & $23.710\,(\pm0.033)$ & $23.921\;(\pm0.022)$ & $7.427\;(\pm0.091)$ \\
    (i) $FTT_{def}$ & $23.780\,(\pm0.090)$ & $23.939\;(\pm0.053)$ & $6.129\;(\pm0.146)$ \\
    (j) $CAFTT_{def}$ & $23.715\,(\pm0.047)$ & $23.807\;(\pm0.017)$ & $6.623\;(\pm0.047)$ \\
    (k) $LocalGLMftt_{def}$ & $23.721\,(\pm0.059)$ & $23.880\;(\pm0.016)$ & $6.832\;(\pm0.099)$ \\
    \botrule
    \end{tabular*}
    \footnotetext{Average and standard deviation of model result 
    over 15 fits with different seeds.}
    \end{table}
\noindent
We can see that there is for all models a clear improvement 
in the loss compared to the mean model. 
The results for the GLMs $(b)\text{-}(d)$ are the same as in Table 5.5 of the book~\cite{wuthrich_statistical_2023}.
Our results for the FNN-based models $(e),(f)$, and $(h)$ closely align 
with those presented in~\cite{richman_localglmnet_2023}. This agreement is supported by the presented standard deviations.
Especially for the transformer based model  $(i)\text{-}(k)$ we can observe that the average predicted claims frequency $avg(\hat{y})$ 
on the test dataset is in comparison to the Poisson GLMs substantially far away from the true mean of $7.35\%$. This is a well-known problem and has to be taken into account when using machine learning models in practice. 
We refer here to~\cite{wuthrich_balance_2022} for a detailed discussion and solutions to this issue. 
By applying a very simple rebalancing factor, which is calculated just on the training data, we can already observe a clear improvement 
for the transformer based models $(i)\text{-}(k)$, see here fore table~\ref{tab:model_rebalanced_results}.
\\ 
\begin{table}[h]
    \caption{Rebalanced model results}\label{tab:model_rebalanced_results}
    \begin{tabular*}{\textwidth}{@{\extracolsep\fill}lrlr}
    \toprule%
    model                   & Train-loss in \%   & Test-loss in \%           & $avg(\hat{y})$ in \% \\
    \midrule
    (e) $Rebalanced:\,FFN_{OHE}$ & $23.752\,(\pm0.033)$ & $23.864\;(\pm0.015)$ & $7.403\;(\pm0.007)$ \\
    (f) $Rebalanced:\,FNN_{EMB}$ & $23.767\,(\pm0.016)$ & $23.826\;(\pm0.015)$ & $7.409\;(\pm0.005)$ \\
    (g) $Rebalanced:\,CANN$ & $23.741\,(\pm0.061)$ & $23.809\;(\pm0.033)$ & $7.405\;(\pm0.006)$ \\
    (h) $Rebalanced:\,LocalGLMnet$ & $23.709\,(\pm0.034)$ & $23.920\;(\pm0.022)$ & $7.407\;(\pm0.005)$ \\
    (i) $Rebalanced:\,FTT_{def}$ & $23.652\,(\pm0.064)$ & $23.815\;(\pm0.036)$ & $7.381\;(\pm0.011)$ \\
    (j) $Rebalanced:\,CAFTT_{def}$ & $23.669\,(\pm0.046)$ & $23.766\;(\pm0.017)$ & $7.392\;(\pm0.005)$ \\
    (k) $Rebalanced:\,LocalGLMftt_{def}$ & $23.696\,(\pm0.066)$ & $23.859\;(\pm0.017)$ & $7.408\;(\pm0.007)$ \\
    \botrule
    \end{tabular*}
    \footnotetext{Average and standard deviation of rebalanced model result
    over 15 fits with different seeds.}
    \end{table}\vfill\eject
    \noindent
At this stage, we already see that the transformer-based models ($FTT_{def}$, $CAFTT_{def}$, $LocalGLMftt_{def}$) clearly outperform their FNN-based counterparts ($FNN_{EMB}$,$CANN$, $LocalGLMnet$). But despite the very long training time of the transformer-based models $(i)\text{-}(k)$, the performance is still moderate. 
This can be traced back to the fact that we didn't perform any hyperparameter tuning and used the default hyperparameter settings.
As mentioned in section~\ref{sec_implementation} the authors of the paper~\cite{gorishniy_revisiting_2021}
showed that an ensemble of already 5 models delivers very similar results as a single hyperparameter-tuned model.
So, we will approximate the real strength via a 5-model ensemble shown in table~\ref{tab:model_ensemble_results}.
\\ 
\begin{table}[h]
    \caption{Average ensemble model results}\label{tab:model_ensemble_results}
    \begin{tabular*}{\textwidth}{@{\extracolsep\fill}lrlr}
    \toprule%
    model                   & Train-loss in \%   & Test-loss in \%           & $avg(\hat{y})$ in \% \\
    \midrule
    (e) $Ensemble:\,FFN_{OHE}$ & $23.715\,(\pm0.022)$ & $23.826\;(\pm0.010)$ & $7.403\;(\pm0.002)$ \\
    (f) $Ensemble:\,FNN_{EMB}$ & $23.743\,(\pm0.003)$ & $23.801\;(\pm0.011)$ & $7.409\;(\pm0.002)$ \\
    (g) $Ensemble:\,CANN$ & $23.701\,(\pm0.043)$ & $23.769\;(\pm0.030)$ & $7.405\;(\pm0.001)$ \\
    (h) $Ensemble:\,LocalGLMnet$ & $23.664\,(\pm0.013)$ & $23.873\;(\pm0.002)$ & $7.407\;(\pm0.002)$ \\
    (i) $Ensemble:\,FTT_{def}$ & $23.592\,(\pm0.003)$ & $23.759\;(\pm0.014)$ & $7.381\;(\pm0.003)$ \\
    (j) $Ensemble:\,CAFTT_{def}$ & $23.630\,(\pm0.018)$ & $23.726\;(\pm0.006)$ & $7.392\;(\pm0.001)$ \\
    (k) $Ensemble:\,LocalGLMftt_{def}$ & $23.645\,(\pm0.034)$ & $23.811\;(\pm0.010)$ & $7.408\;(\pm0.003)$ \\
    \botrule
    \end{tabular*}
    \footnotetext{Average results 
    over 3 ensembles consisting of each 5 rebalanced models.}
    \end{table}
\noindent
So we can see here very clearly that the transformer-based actuarial models $CAFTT$ and $LocalGLMftt$ 
clearly outperform there $FFN$ based counterparts $CANN$ and $LocalGLMnet$ on the testset. Especially the results for the $CAFTT$ are setting, to the best of our knowledge, a new benchmark on this very popular dataset.
\section{Conclusion and Discussion}\label{sec_conclusion}
We showed that with the CAFTT and LocalGLMftt, we can continue to build upon the very established GLMs 
in place but use the predictive power of the transformer. 
Depending on whether better interpretability or predictive power is more important for the task, 
one can choose the preferred models for the task. 
In the case of the LocalGLMftts, we still have the option of an interpretable model, while performing better than the native LocalGLMnet and the $FNN_{OHE}$ model, and if raw performance is needed, the CAFTT is to choose. Note in the case of the ensemble models, the interpretability properties of the LocalGLMftts are still there since the $\beta_{FTT}(x)$ can be aggregated. 
Because we can use categorical features as an input for the FTT, 
we can use the same feature engineering as currently in place at the companies 
but with the predictive power of the transformer. 
So, there is no need for change in the current application pipelines, be it in the frontend or backend.
Because on a big dataset, the transformer model training takes a lot of time, 
it is conceivable for the update process not to re-fit the entire transformer model 
but to take the pre-trained model from the previous year and only train 
the last layers on an updated dataset. 
At this point, it is important to notice that there is still some work to do 
before the GLM models can be fully replaced by machine-learning models 
in actuarial departments around the world.
One factor besides the interpretability, is the so-called time consistency property that 
actuaries are very focused on when building a model. 
This property takes into account that only the time-stable effects 
are modeled and that the model is not overfitting due to the yearly effects 
of which in the insurance world, there have been a lot in the previous years. 
In regards to the interpretability, one has to consider that often, in actuarial departments 
very big GAM models with many interactions are in place, and as~\cite{richman_believing_2019} 
mentions those models are also very hard to interpret. 
But one has to consider that it is very important for actuaries not only to interpret 
the model but also to be able to change the behavior of the model in regards to specific features 
in a controlled way on a very granular level. 
This is important to accommodate inconsistencies in the data as well as current trends 
in technology and customer behavior. 
Although for machine learning models, there are monotonic constraints on the feature level available 
but this brings by far not the flexibility for expert judgment that is currently present for the GLMs.
Furthermore, as \cite{lindholm_multi-task_2023} describes, one has to take measures to avoid indirect discrimination via predictive models. 
Another practical point that we want to mention here is that under certain circumstances, 
the live prediction time for the forward pass of the transformer models needs 
to be optimized for live applications. 
GLMs and GAMs are great methods, which is the reason why they are still the predominant tools in insurance companies. 
By extending the solid foundation laid by~\cite{schelldorfer_nesting_2019} and~\cite{richman_localglmnet_2023}, 
and implementing the innovative methodology of~\cite{gorishniy_revisiting_2021}, 
we aspire to actively contribute to the ongoing process of GLM and GAM improvements.
\vspace{8mm}
\\\text{ }\\
\noindent
\textbf{Acknowledgements:}
This research was supported by Allianz Versicherungs-AG.
The author thanks his colleagues from the actuarial department and his supervisors
Dr. Renata Gebhardt (Allianz Versicherungs-AG) and Prof. Dr. Christian Heumann (LMU) for fruitful discussions and valuable remarks to improve earlier versions of this manuscript.
\vspace{6mm}
\\\text{ }\\
\noindent
\textbf{Funding:}
This manuscript was written during the author's employment as an actuary at Allianz Versicherung-AG.
He was granted to devote part of his working time to research.
\vspace{6mm}
\\\text{ }\\
\noindent
\textbf{Financial interests:}
The author declares they have no financial interests.
The author has no conflicts of interest to declare that are relevant to the content of this article.
\bibliography{my_bibliography}
\end{document}